\documentclass[letterpaper, 10 pt, conference]{ieeeconf}  

\IEEEoverridecommandlockouts                              

\usepackage{amsmath} 
\usepackage{amssymb}  

\usepackage{graphicx} 
\usepackage{subcaption}
\usepackage{multirow} 
\usepackage[table]{xcolor}
\usepackage{makecell}
\usepackage{hyperref}
\usepackage{algorithm}
\usepackage{algcompatible}
\usepackage{bm}

\title{\LARGE \bf
Industrial Robot Grasping with Deep Learning\\ using a Programmable Logic Controller (PLC)}

\author{Eugen Solowjow$^{1}$, Ines Ugalde$^{1}$, Yash Shahapurkar$^{1}$, Juan Aparicio$^{1}$,\\ Jeff Mahler$^{2,3}$, Vishal Satish$^{2}$, Ken Goldberg$^{2}$, Heiko Claussen$^{1}$
\thanks{$^{1}$Siemens
        {\tt\small \{eugen.solowjow, ines.ugalde, yash.shahapurkar, juan.aparicio, heiko.claussen\}@siemens.com}}%
\thanks{$^{2}$AUTOLAB at University of California, Berkeley{\tt\small \{jmahler, vsatish, goldberg\}@berkeley.edu}}%
\thanks{$^{3}$Ambidextrous Laboratories, Inc }%
}

\begin{document}

\maketitle
\thispagestyle{empty}
\pagestyle{empty}

\begin{abstract}
Universal grasping of a diverse range of previously unseen objects from heaps is a grand challenge in e-commerce order fulfillment, manufacturing, and home service robotics.
Recently, deep learning based grasping approaches have demonstrated results that make them increasingly interesting for industrial deployments. This paper explores the problem from an automation systems point-of-view. 
We develop a robotics grasping system using Dex-Net, which is fully integrated at the controller level. 
Two neural networks are deployed on a novel industrial AI hardware acceleration module close to a PLC with a power footprint of less than 10 W for the overall system. 
The software is tightly integrated with the hardware allowing for fast and efficient data processing and real-time communication.
The success rate of grasping an object form a bin is up to 95\% with more than 350 picks per hour, if object and receptive bins are in close proximity.
The system was presented at the Hannover Fair 2019 (world's largest industrial trade fair) and other events, where it performed over 5,000 grasps per event.
\end{abstract}
\section{INTRODUCTION}
Universal reliable robot grasping of a diverse range of objects is a challenging task.
The challenges arise from imprecisions and uncertainties in sensing and actuation. 
A solution to universal grasping will enable automation of many industrial tasks that are mostly performed by humans today such as e-commerce order fulfillment, manufacturing and home service robotics.
Recently, deep learning based approaches to universal grasping demonstrated progress in terms of accuracy, reliability and cycle times, showing promise for industrial deployment.
However, designing a flexible automation system that has deep learning at its core is from an automation systems point of view a challenging undertaking, which received little attention from the robotics community so far.

Robot control systems in today’s industrial environments give robots the ability to follow predefined trajectories.
These systems are considered mature and ten thousands of them have been deployed across factories in different industries. 
However, most robots in factories lack the ability to handle variability and uncertainty, which are key elements in universal grasping.
On the other hand, robot control systems based on machine learning methods have not yet been widely adopted in factories.
They are mainly situated in research environments where the focus is on algorithmic innovation.  
\begin{figure}[hbtp]
      \centering
\begin{subfigure}{1.0\linewidth}
    \centering
    \includegraphics[width=1.0\linewidth]{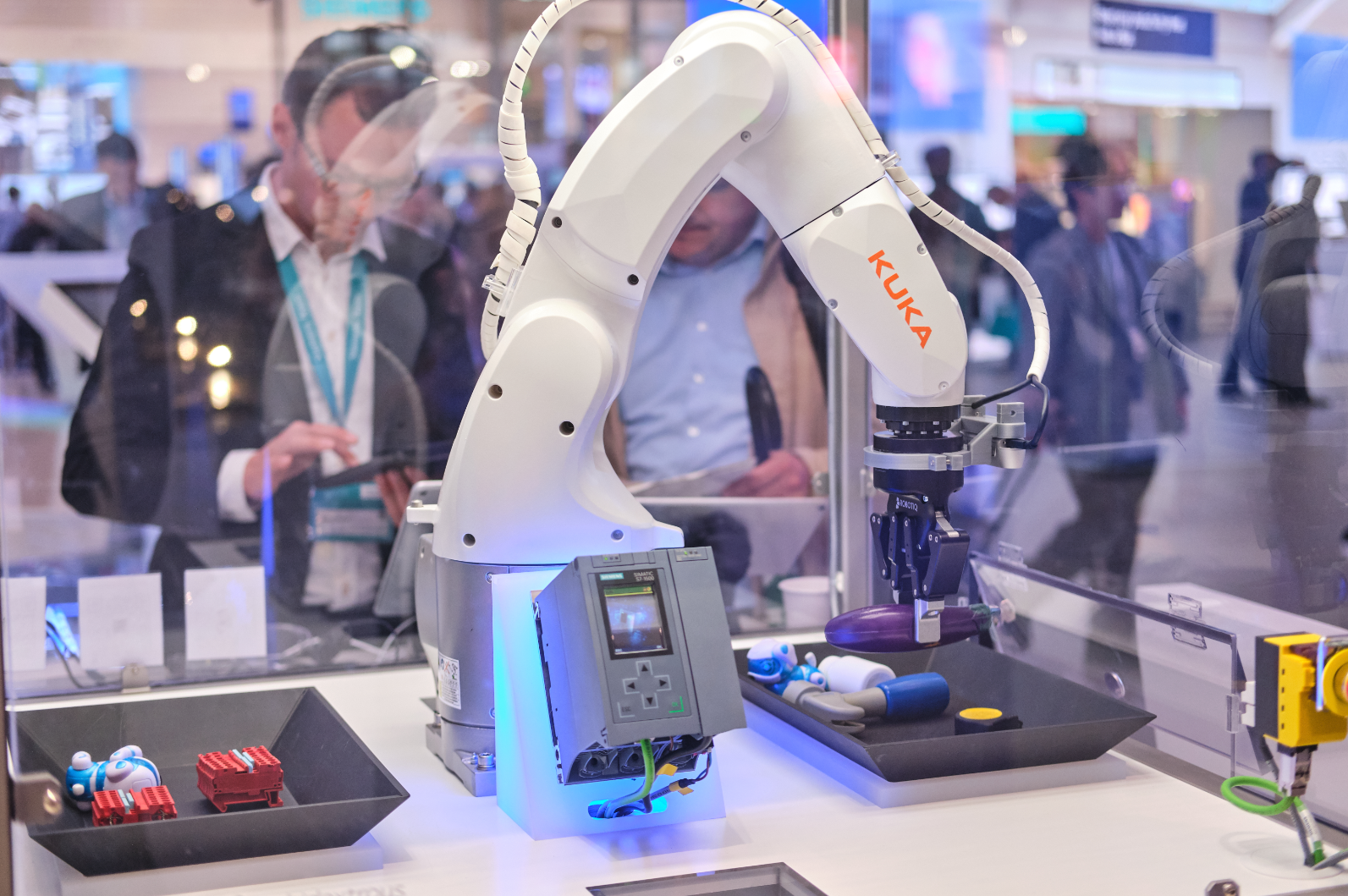}
\end{subfigure}
  \caption{Industrial system for universal grasping at Hannover Fair 2019, Germany.}
      \label{fig:system_at_hmi}
\end{figure}
With the increasing interest in deep learning based automation, the question arises how these algorithmic progresses can be adopted by industry and integrated into the existing automation landscape \cite{luo18, johannink19}.
Conventional automation typically assumes full predictability of the world the system deals with.
This assumption does not hold true for universal grasping with its shear endless number of object constellations, which requires automating of the unpredictable.
A paradigm shift is required from traditional automation systems to machine learning based automation systems.

In this paper we describe an industry-compatible robotic system for universal grasping enabled by deep learning. 
We study the problem from an automation systems point of view: “How to design a deep learning based automation system that is tightly integrated with current automation paradigms such as PLC control and can be deployed in an industrial production?” 
The resulting system was exhibited at Hannover Fair 2019 (world’s largest industrial trade fair), where it ran for five consecutive days and performed over 5,000 grasps, see Fig. \ref{fig:system_at_hmi}
Since its debut it has been replicated and has been shown at various other events.

\section{PROBLEM DEFINITION AND REQUIREMENTS}
\begin{figure}[hbtp]
      \centering
\begin{subfigure}{1.0\linewidth}
    \centering
    \includegraphics[width=1.0\linewidth]{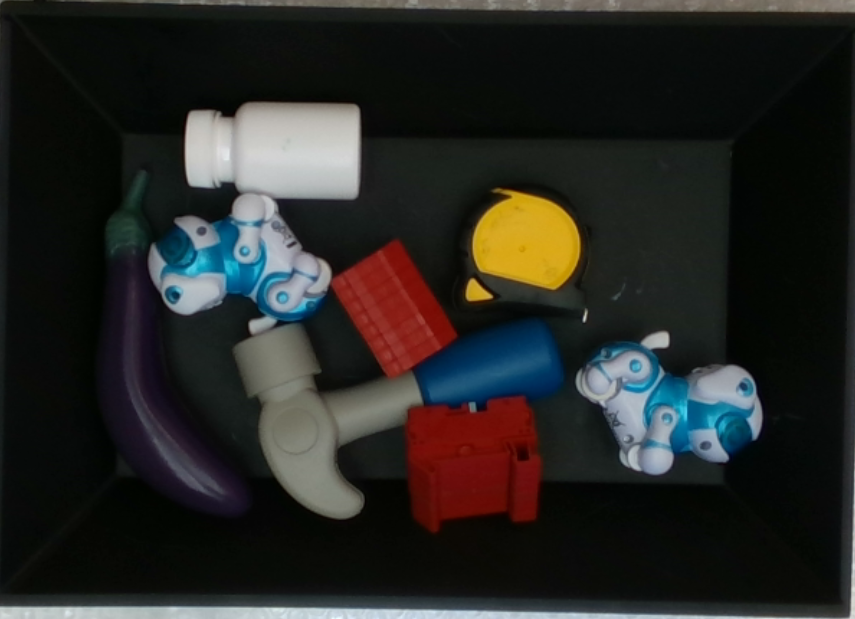}
\end{subfigure}
  \caption{Sample objects for grasping in bin.}
      \label{fig:sample_bin}
\end{figure}

The problem of universal grasping gives robots the ability to pick previously not encountered objects in arbitrary constellations.
It can be defined by the constellation of the objects (e.g. singulated or heaps), the type of hardware (e.\,g. single or dual robot arms), as well as on the type of end-effector (e.g. vacuum gripper, two finger parallel jaw gripper, or custom gripper designs). 
Moreover, the task description can vary.
For example, clearing a bin of objects requires a different approach than picking one desired object from a heap.
The problem addressed in this paper is as follows. 
A bin of objects is presented to the system as shown in Fig. \ref{fig:sample_bin}
A user selects through a Human Machine Interface (HMI) the objects to be picked from the bin and the amount of requested objects.
The robot picks the requested objects from the bin and places them into another bin.
The system informs the user, if any of the requested objects are not available.
\section{RELATED WORK}
Universal robotic grasping has been widely studied during the last decades, and, until today, it is still considered a challenging problem. Initiatives, like the Amazon Robotic Challenge aim at pushing the state of the bin picking research towards production \cite{c0, c1, c2}. The difficulty of universal grasping lies in the great variability of bin picking scenarios, such as heterogeneous, potentially unknown objects, which may be arbitrarily positioned in presence of occlusions, super-positions, hindered reachability, etc. Early approaches computed grasps on known 3D models of objects using analytic methods and planned grasps online by matching sensor data to known objects. However, these approaches proved susceptible to uncertainty in sensing and could not generalize well to novel objects. Recent learning-based approaches have leveraged deep neural networks trained on large amounts of data that can quickly generalize to novel objects. Empirical approaches to collecting this data have proven to be time-consuming and prone to mislabeled data. An alternative promising approach is to rapidly generate massive synthetic datasets using analytic metrics and domain randomization for robust sim-to-real transfer.

A large number of object recognition bin picking solutions follow a two-step approach: object detection and pose estimation followed by model-based grasp planning. Typically, Convolutional Neural Networks (CNNs) are employed in the object recognition task \cite{c1, c2} to provide either bounding boxes or segmentations of objects of interest. Other approaches \cite{c3, c4} build on this pipeline to integrate pose correction in the robot motion stage. A representative example of the two-step approach is presented in \cite{c2} where a 3D segmentation is obtained by projecting and merging individual RGB-D segmentations from multiple cameras positioned to capture the whole scene. Pose estimation is achieved by fitting given 3D models of the objects to the 3D segmented point cloud. A series of heuristics is then applied to calculate optimal grasping points. This sophisticated approach leads to a total perception time of 15-20 seconds on a workstation equipped with large-scale processors (such as an Intel E2-1241 CPU 3.5 GHz and an NVIDIA GTX 1080). 

Previous work often assumed given 3D models of objects, and some even labeling of training data for the segmentation CNNs. These approaches may prove impractical for scalable applications which continuously deal with novel objects for which limited data is available, as typically seen in intralogistics and warehousing scenarios. As an alternative, object-agnostic data-driven grasping has been studied in \cite{c5, c6, Bohg_c7_0, Kappler_c7_1, Morrison_c7_2, Mahler17_c7_3, c7}. In \cite{c7} the authors present Dex-Net 4.0, where bin picking is formalized as a POMDP problem to enable simulate and learn from synthetic data (synthesizing depth images of over 5000 unique object heaps) robust grasping policies for parallel-jaw and suction grippers. Experiments on 50 novel objects using the high-end Photoneo Phoxi 3D camera suggest that Dex-Net 4.0 generalizes to new objects. Similar to other approaches, this system was deployed on a large-scale processing workstation (equipped with quad-core Intel i7-6700 and NVIDIA TITAN Xp).

Computer vision bin picking solutions can be found in industry as well, however there is reduced information as to what algorithms or principles are applied in the search for grasping points. iRVision \cite{c8} is FANUC's visual detection system which uses 3D vision (structured light mapping and laser projection) to achieve some form of detection and pose estimation of workpieces in aid of different manufacturing processes, such as bin picking. Some algorithms used by the manufacturer are Geometric Pattern Matching and Blob Detection; it is then understood that substantial prior information about the objects is required.

\section{MECHATRONIC SYSTEM DESIGN}
\begin{figure}[hbtp]
      \centering
\begin{subfigure}{0.95\linewidth}
    \centering
    \includegraphics[width=1.0\linewidth]{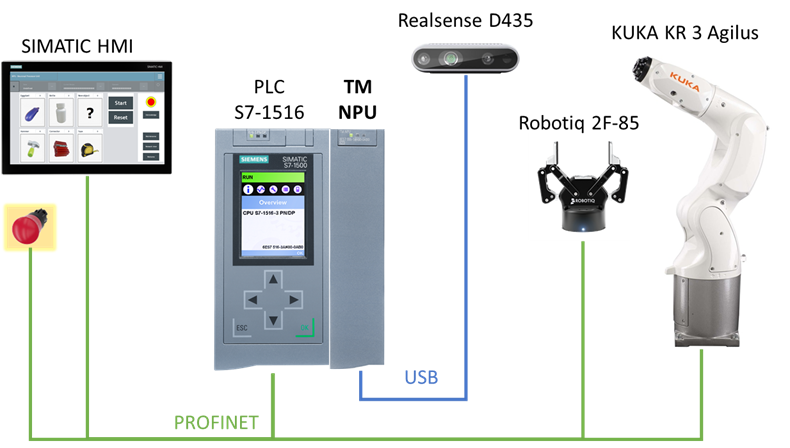}
\end{subfigure}
  \caption{The screen (Human-Machine-Interface) displays a list of objects for the user to select. The PLC receives request from the HMI and consequently triggers the algorithm in the TM Neural Processing Unit (TM NPU). Once a grasp is found, the PLC actuates both robot and gripper to execute the grasp.}
      \label{fig:system_design}
\end{figure}

The system design of the industrial bin picking solution is centered around an industrial Programmable Logic Controller (PLC), a Siemens S7-1516. 
PLCs are high-reliability automation controllers suitable for harsh environments that provide hard-realtime capabilities. 
Choosing the computational components at the controller level is in contrast to most other work, where typically PC workstations are used. The advantages are the highly embedded nature of the solution, the small power footprint and the seamless integration into the existing automation ecosystem. In Fig. \ref{fig:system_design} the PLC can be seen as the master of the system; it orchestrates the interactions among all the devices. A Human Machine Interface (HMI) touch panel was integrated with the PLC program to present an array of objects which the user can select to be picked. The HMI also allowed maintenance-mode control of the robotic elements and showed diagnostic information of all connected elements. An E-Stop push button was also integrated with the PLC; its activation halts the entire system. Both HMI and E-stop are connected using industrial standard communication PROFINET to the PLC. The mechanical components of our bin picking solution comprise the KUKA KR 3 Agilus 6-DOF robotic arm and the Robotiq 2F-85 parallel jaw gripper. The KUKA KR 3 and controller (KR C4 Compact) are connected via PROFINET to the PLC, where seamless control over the robot is enabled with the KUKA PLC mxAutomation package \cite{c9}. An Intel RealSense D435 RGB-D camera serves as the sensory input to the system. It is mounted on the robot wrist and connects via USB3. The D435 is widely available and offers a low price point.

The Technology Module Neural Processing Unit (TM NPU)\footnote{www.siemens.com/tm-npu} in Fig. \ref{fig:system_design} is dedicated to Deep Learning and allows PLC-based automation systems to incorporate efficient Neural Network computations. The TM NPU is equipped with an Intel MyriadX SoC, which has two LEON CPUs and 16 SHAVE parallel processors capable of accelerating neural networks in hardware with a compute capacity of up to 4 TOPs. The TM NPU couples with the PLC using Siemens S7 backplane communication \cite{c10} by which they share information in real-time. The NPU algorithms is invoked by the PLC, upon HMI user requests. The TM NPU uses RGB-D inputs from the RealSense D435. The algorithm returns pixel coordinates for grasping as well as object identity. The PLC transforms the coordinates to the robot frame and commands the robot motion; at the same time, the identified object is highlighted in the HMI.


\section{SOFTWARE DESIGN} 
%
The challenges in designing the software arise from the required flexibility and the unpredictability of the grasping scenario as described in Sec. II.
The system comprises of two main computational entities: A PLC and the TM NPU. 
The architecture of the control system has to reflect the use case flexibility by means of Deep Learning and at the same time to preserve the benefits of an industrial automation system, namely real-time properties, robustness, and safety.

\begin{figure}[hbtp]
      \centering
\begin{subfigure}{1.0\linewidth}
    \centering
    \includegraphics[width=1.0\linewidth]{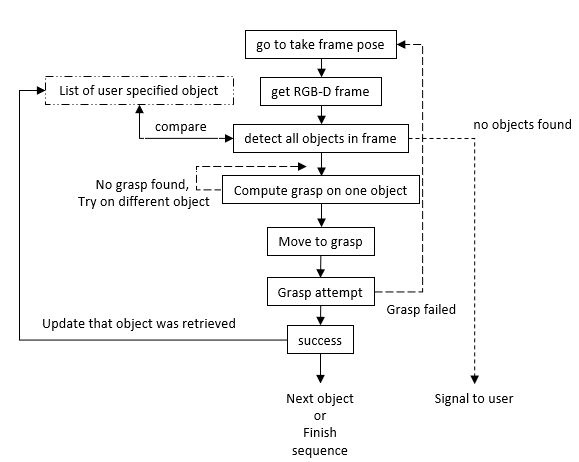}
\end{subfigure}
  \caption{Simplified Deterministic Finite Automaton (DFA) for processing of one RGB-D frame.}
      \label{fig:automaton}
\end{figure}

The desired system behavior can be described with states, between which the system transitions based on events, and their relationships.
We use a Deterministic Finite Automaton (DFA) to model the desired system behavior, which is illustrated in Fig. \ref{fig:automaton} for one RGB-D camera frame.
The automaton describes the actions and states for a successful grasp, but also actions for common failures such as “no object grasped”.
The system maintains a list of user desired objects, which is provided through the HMI.
The automaton is repeatedly traversed until the list has been either fulfilled or none of the requested objects can be detected in the bin.

\subsection{SOFTWARE ARCHITECTURE AND MODULES}

\begin{figure*}[hbtp]
      \centering
    \includegraphics[width=1.0\linewidth]{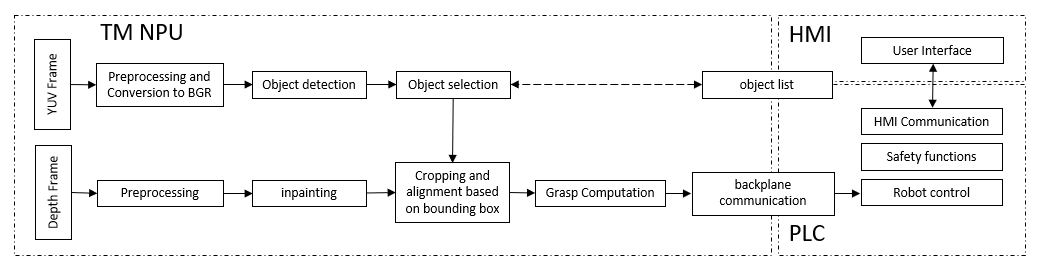}
  \caption{Main software modules distributed amongst TM Neural Processing Unit (TM NPU), Human-Machine-Interface (HMI) and Programmable Logic Controller (PLC).}
      \label{fig:software_modules}
\end{figure*}

Following the hybrid automaton of the previous subsection, we can derive the key software modules, which enable the system to react to events and cope with the flexibility of the usecase. 
Figure \ref{fig:software_modules} illustrates the main software modules and their relations.
All software modules related to image processing including the grasp computation are implemented on the TM NPU, whereas robot control, the interfacing with the HMI and safety functions are implemented on the PLC.
The TM NPU and PLC communicate through a backplane, while the PLC and the HMI communicate through Ethernet, which is a typical setup for this type of automation hardware.

In the remainder of this subsection we describe the key functionalities of the main software modules.
All software modules that run on the TM NPU are implemented in C/C++ on top of the real-time operating system RTEMS\footnote{www.rtems.org}.
The RGB-D input to the TM NPU is an RGB-D camera (in this case an Intel RealSense D435), which provides streams of color and depth frames.

\textbf{Preprocessing and Conversion (color frame):} Camera interfaces provide frames in a variety of different encodings such as YCbCr422i. Depending on the encoding the color frames need to go through several basic image processing steps such as scaling, cropping, de-interleaving and conversion to BGR or RGB, which are the most common formats for neural network inputs. These image processing steps are efficiently implemented and parallelized on the SHAVE processors of the TM NPU MyriadX SoC.

\textbf{Object Detection (color frame):} The preprocessed color frames are fed into an object detection neural network. Popular neural networks are MobileNet-SSD, YOLO, and Faster RCNN.
The output of an object detection algorithm are bounding boxes with corresponding labels and the classification score. The object detection neural network is described in the following Subsection.

\textbf{Preprocessing (depth frame):} Depth information is usually encoded as a single channel 2 byte stream. The preprocessing consists of scaling and cropping operations.

\textbf{Inpainting (depth frame):} Depth frames have often missing data when surfaces are too glossy, thin, bright, far or close from the camera. Depth inpainting fills these “holes”. With increasing popularity of commodity-grade depth cameras (e.g. Intel RealSense, Microsoft Kinect) inpainting algorithms received a lot of attention in recent years. We implemented a custom inpainting approach that is based on \cite{inpaint_c11}.

\textbf{Object Selection and Bounding Box based Cropping:} The detected objects are compared with the user selected objects and a matching is determined. For all detected objects that are also part of the user list, a pairwise bounding box overlapping is computed. The object that overlaps least with other bounding boxes is chosen for grasping. This object is cropped out in the depth frame. Note this is a decisive step, which couples object detection with grasping and where the decision is made which object to grasp. Other methods can be used to choose the next best object or even a unified approach of detection and grasping can be deployed.

\textbf{Grasp Computation:} The grasp is computed in this deployment with an FC-GQ-CNN \cite{Satish19_c13}. The cropped depth image is fed into the network, which outputs a tensor representing the grasps and their quality score. The grasp neural network is covered in Subsection V. C.

\textbf{PLC and HMI:} The PLC contains the main state machine and orchestrates the overall process. It is programmed with ladder logic. The PLC receives the user specified object list from the HMI, tasks the NPU to compute grasps against the current active object list, executes the robot and gripper motions and reports back to the HMI to keep the user informed. 

\subsection{OBJECT DETECTION}
The object detection algorithm inputs color images. It outputs bounding boxes for each object in the image, the class label associated with each bounding box and the confidence score associated with each bounding box and label. Since the advent of deep learning, neural network based object detection approaches have shown to be the most accurate algorithms. In recent years, MobileNet SSD (Single Shot Multibox Detection) has proven to be an efficient Convolution Neural Network architecture targeted towards mobile and embedded vision applications. The classification base network is thereby MobileNet, which is pretrained on ImageNet for set of discerning, discriminating filters. By performing depth wise separable convolutions, MobileNet allows a lesser number of tunable parameters which results in light weight deep neural networks. The SSD part of the object detection pipeline discretizes the output space of bounding boxes into a set of default boxes over different aspect ratios. At prediction time, the network generates scores for the presence of each object category in each default box and produces adjustments to the box to better match the object shape.
MobileNet SSD has to be trained on a dataset that is recorded prior to deployment. The dataset contains images of the objects that the system is supposed to recognize for grasping at runtime. We use 200 training images for sufficient performance.
\subsection{GRASP COMPUTATION}
Once an object is chosen, a robust grasp must be quickly planned in order to transport the object to the collection bin. 
\begin{figure*}[hbtp]
      \centering
    \includegraphics[width=1.0\linewidth]{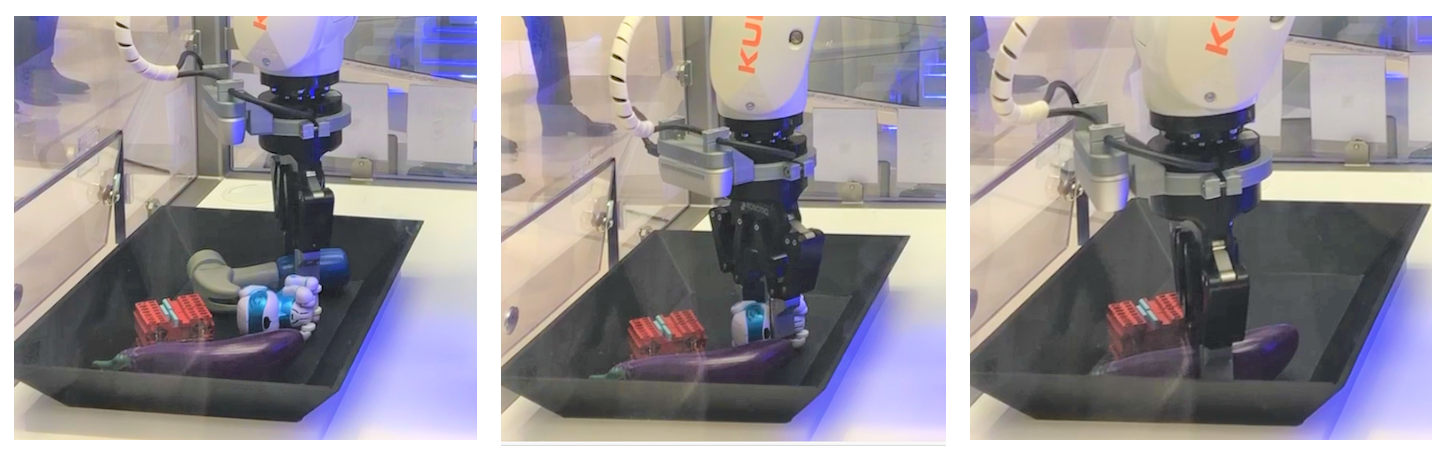}
  \caption{Snapshots of a representative bin picking operations during Hannover Fair 2019.}
      \label{fig:res_sequence}
\end{figure*}
The system in this paper utilizes an extension of Dex-Net 4.0, the Fully Convolutional Grasp Quality Convolutional Neural Network (FC-GQ-CNN), which given an input point cloud rapidly evaluates millions of four degrees-of-freedom (3D position and planar orientation) grasps in a single forward pass of the network and chooses the highest quality one for execution. The network is trained on synthetic data using the methodology of Dex-Net 4.0 with a Robotiq 2F-85 parallel jaw gripper and an Intel RealSense D435 depth camera. In order to model the noise in the RealSense camera, the images are augmented with synthetic noise sampled from a Gaussian Process \cite{Mahler17_c7_3} to reflect the significant levels of noise compared to the Photoneo PhoXi S sensor that is used in Dex-Net 4.0. After training, the network was compiled for the MyriadX SoC and deployed on the TM NPU. While the neural network is not trained using any of the demo objects, it generalizes and is able to grasp them as noted by the performance in Section VI.

\section{EVALUATION}
The robotic system was integrated into a custom robot cell of 1\,m$\times$1\,m footprint so that it can be shipped and deployed at exhibitions and fairs worldwide. The cell was equipped with two bins of size 45\,m$\times$25\,m$\times$8\,m from which the robot can pick and place objects. 
The object detector was trained for six different objects, which are shown in Fig. \ref{fig:res_obj_det}. Note that while the object detector cannot recognize objects that it was not trained for, the FC-GQ-CNN is  capable of grasping any previously unseen object. 
One of the requirements for the system was ease of use. 
The robot cell would be displayed at various exhibitions world-wide, where local personal (usually neither robotics nor Deep Learning experts) should be able to setup and run the demo within short time.
We implemented an easy-to-use user and debugging interface so that within few minutes anyone can run the flexible grasping system.

\subsection{Performance}
The bin picking system was introduced at Hannover Fair 2019 in Germany. The system ran for five consecutive days, eight hours each day. It performed approximately 1500 grasps per day. The workflow usually consisted of packing the input bin with objects and then selecting desired objects on the HMI screen for the robot to pick. Figure \ref{fig:res_sequence} shows snapshots for a representative grasping task, where the robot was tasked to get the hammer, the dog, and the eggplant.
The grasping accuracy of the system is in general dependant on the amount of objects in the bin. The denser the bin is packed, the more difficulties the system experiences to retrieve the objects of interest, in particular if the selected objects are not easily reachable. For lightly packed bins, where each object allows for a grasp as shown in Fig. \ref{fig:sample_bin}, the success rate of grasping an object from the bin was ca. 90\%-95\%. A detailed analysis of failure modes is discussed in the subsequent subsection. 

As expected, the system's the computational times and power footprint are its strengths.
The tight integration of HMI, PLC, TM NPU and robot control results in a total processing time of less than 1\,sec from sending the object request to the PLC until the start of the robot motion. During this time the computations detailed in Figs. \ref{fig:automaton} and \ref{fig:software_modules} are performed, which includes execution of two deep neural networks. The inference times of  MobileNet SSD and FC-GQ-CNN are 350\,ms and 70\,ms respectively. 
The system performs 200-250 picks per hour. However, the receptive bin is placed at the opposite side of the robot cell resulting in the maximum travel distance for the robot arm to drop off a picked object. If the bins are placed in close proximity the system achieves over 350 picks per hour.
The combined power consumption of PLC and TM NPU is less than 10\,W.

\begin{figure}[hbtp]
      \centering
    \includegraphics[width=0.9\linewidth]{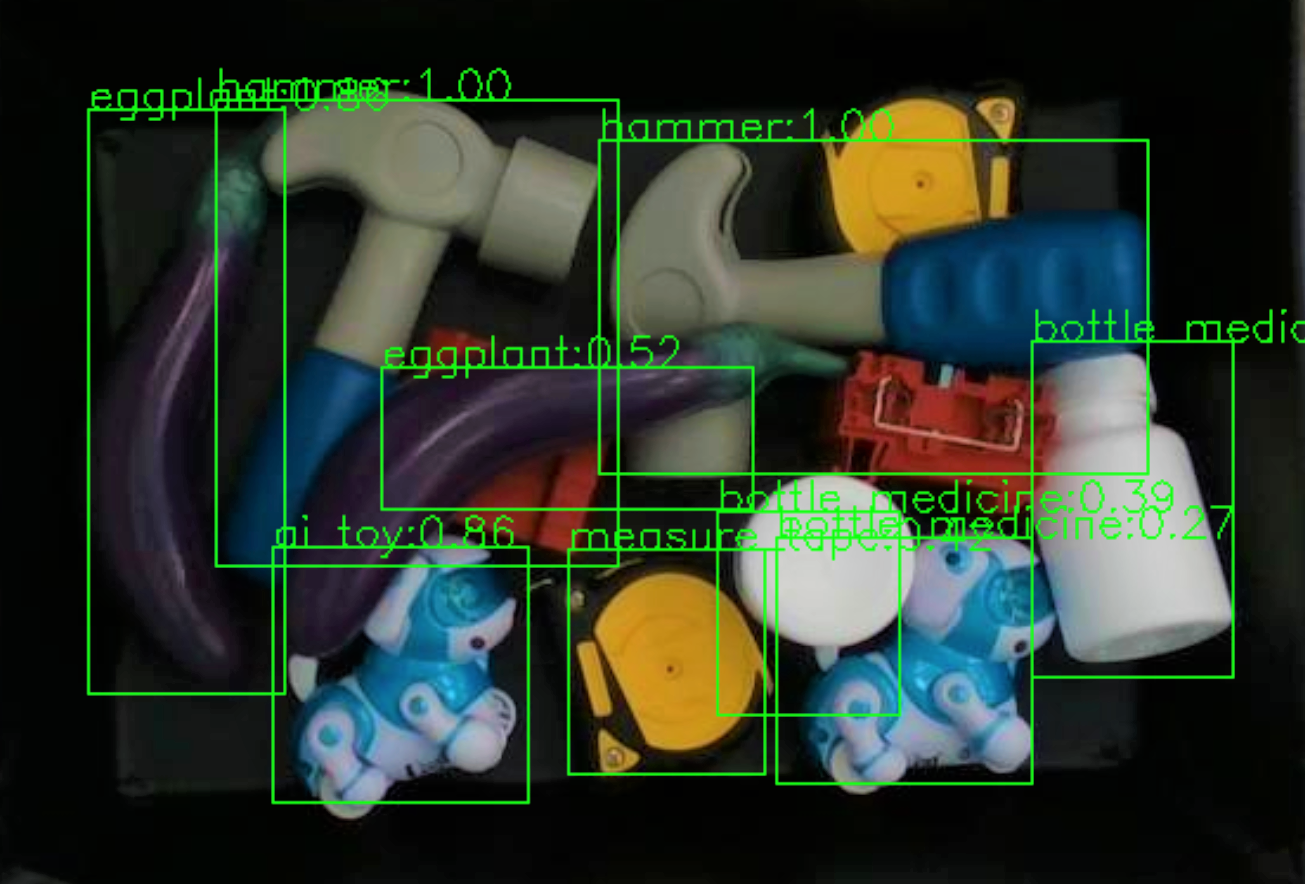}
  \caption{Object detection results with confidence levels computed with MobileNet SSD.}
      \label{fig:res_obj_det}
\end{figure}

Figure \ref{fig:res_obj_det} illustrates the output of the object detection pipeline with MobileNet SSD. 
While most objects are recognized reliably, note that some objects are not detected at all. In our case this is usually not a problem, because the objects that are not detected are usually covered objects lying at the bottom of the bin, which would not be grasped at this instance. Since object detection is repeated after every grasp, the covered objects will be "discovered" after the top objects are cleared. However, if the user selects the few objects that are the bottom of the bin, the system can miss them.

\begin{figure}[H]
      \centering
\begin{subfigure}{0.45\linewidth}
    \centering
    \includegraphics[width=1.0\linewidth]{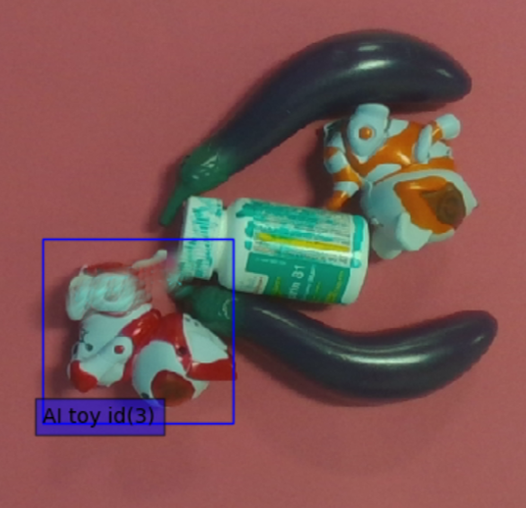}
    \caption{Color frame with bounding box.}
      \label{fig:res_grasp_a}
\end{subfigure}
\hfill
\begin{subfigure}{0.45\linewidth}
    \centering
    \includegraphics[width=1.0\linewidth]{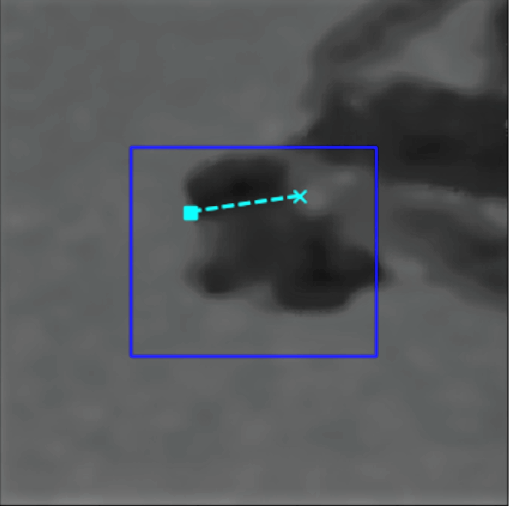}
    \caption{Depth frame with computed grasp.}
      \label{fig:res_grasp_b}
\end{subfigure}
   \caption{Sample grasp computation result, where toy animal was selected for grasping.}
      \label{fig:res_grasp}
\end{figure}

A representative grasp computation output can be seen in Fig. \ref{fig:res_grasp}. After a bounding box provided by the object detector has been selected in the color frame as seen in Subfig. \ref{fig:res_grasp_a}, the depth frame is aligned to the color frame. Next, the cropped and centered depth image is processed by the FC-GQ-CNN, which outputs grasp coordinates as shown in Subfig. \ref{fig:res_grasp_b}. Only grasps within the bounding box are considered.

\begin{figure}[H]
      \centering
\begin{subfigure}{0.45\linewidth}
    \centering
    \includegraphics[width=1.0\linewidth]{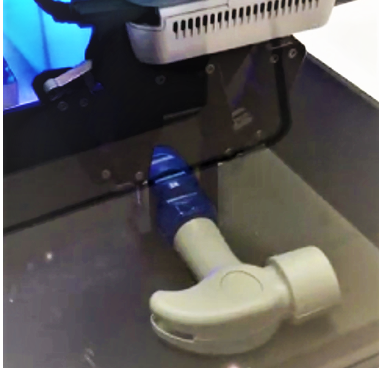}
    \caption{No obstacles.}
      \label{fig:res_hammer_a}
\end{subfigure}
\hfill
\begin{subfigure}{0.45\linewidth}
    \centering
    \includegraphics[width=1.0\linewidth]{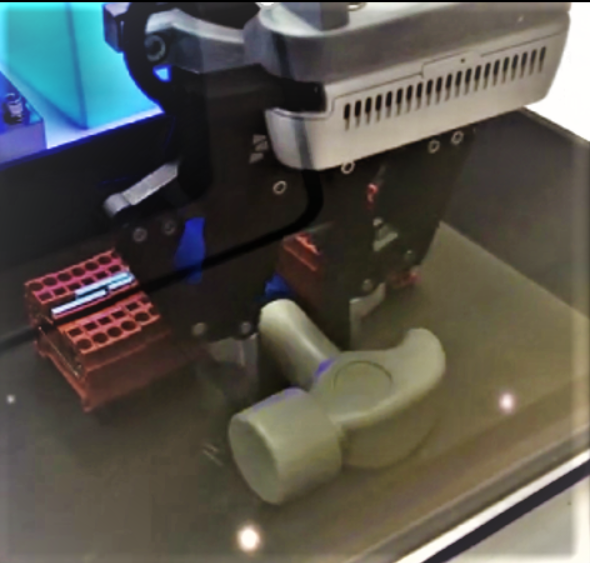}
    \caption{Handle obstructed.}
      \label{fig:res_hammer_b}
\end{subfigure}
   \caption{On the fly adjustment of grasp computation by FC-GQ-CNN without explicit behaviour programming.}
      \label{fig:res_hammer}
\end{figure}

One of the advantages of deep learning based approaches is the capability to adapt to variations in the environment. This behaviour is demonstrated in Fig. \ref{fig:res_hammer}. 
A hammer is presented to the system for grasping. The system generally picks the hammer by the handle as shown in Subfig. \ref{fig:res_hammer_a}. If the handle is obstructed by obstacles, so that the system cannot grasp the hammer by the handle anymore, FC-GQ-CNN computes other grasp points in order to avoid collisions with the obstacles, see Subfig. \ref{fig:res_hammer_b}. This logic is not explicitly programmed, but emerges from the synthetic training samples that the neural network used to "learn" grasping. 
\subsection{Failure Modes}
\begin{figure}[hbtp]
      \centering
      \begin{subfigure}{0.45\linewidth}
    \centering
    \includegraphics[width=1.0\linewidth]{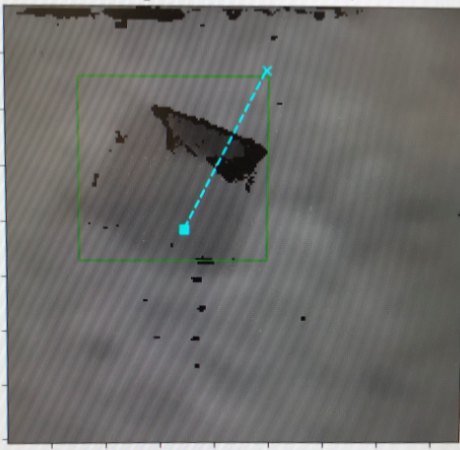}
    \caption{FC-GQ-CNN computes a grasp on bulge that is caused by missing information in the depth image.}
      \label{fig:failure_inpaint}
\end{subfigure}
\hfill
\begin{subfigure}{0.45\linewidth}
    \centering
    \includegraphics[width=1.0\linewidth]{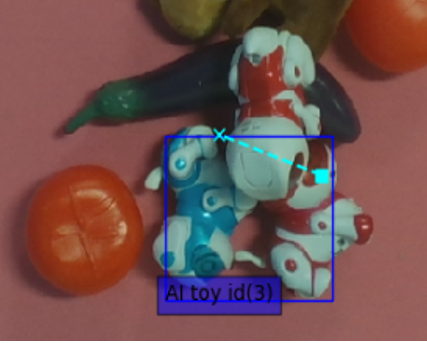}
    \caption{Object detection provides incorrect bounding box.}
      \label{fig:failure_obj_det}
\end{subfigure}

   \caption{Failure modes in object detection and grasp computation.}
      \label{fig:failure_cases}
\end{figure}

During the Hannover Fair 2019 the system performed more than 5000 grasps. This gives us the opportunity to study common failure modes.
The sources of grasping failures are due to sensor modalities, object detection output, grasp computation output or grasp execution. 
Two failure cases due to object detection and grasp computation are shown in Fig.~\ref{fig:failure_cases}.
In both subfigures the respective bounding box of the selected object and the grasp axis (in cyan) are shown. The depth frames of consumer-grade RGB-D cameras often have missing data. While inpainting algorithms can compensate for some of the missing information, they are also prone to create artifacts in the depth image. These artifacts can appear as a bulge part of the object. If such depth images are fed into the FC-GQ-CNN, grasps can be computed on these bulges. 
In Subfig. \ref{fig:failure_inpaint} the object is flat, but the depth image shows a bulge and FC-GQ-CNN computes a grasp there.
A common failure mode in object detection, besides objects not being detected at all, is the wrong size or number of bounding boxes, when objects of the same class cluster together as seen in Subfig. \ref{fig:failure_obj_det}.

\section{CONCLUSION}
In this paper we presented an industrial robotic system for universal grasping. The system uses deep learning and was implemented as a PLC-based automation system. We demonstrated the system during Hannover Fair 2019, where it ran for five consecutive days and performed more than 5,000 grasps. The tight integration of HMI, PLC, TM NPU and robot control allowed for  fast data processing. The total computation time was less than 1\,sec from sending the object request to the PLC until the robot starts moving towards the object, which includes execution of two deep neural networks.  
The system performed 200-250 picks per hour, whereby the receptive bin was placed at the opposite side of the robot cell resulting in the maximum travel distance for the robot arm. For  bins placed in close proximity the system achieved over 350 picks per hour.
The combined power consumption of PLC and TM NPU was less than 10\,W.
The system consisted of off-the-shelf automation components. 
One of the limitations of the presented system was due to depth sensing. Consumer grade RGB-D cameras, as the one used in the presented system, provide more noisy depth frames than industrial grade depth scanners. However, the latter are significantly more expensive rendering them often as uneconomic solutions.

In our future work, we will extend the functionality to other end-effectors such as vacuum grippers to give the system more flexibility. We will explore semantic segmentation instead of object detection, because it leads to exact shapes of the object instances as opposed to bounding boxes. The dataset creation for semantic segmentation will adopt the methodology presented in \cite{danielczuk17_c12}.


\begin{thebibliography}{99}


\bibitem{luo18} J. Luo, E. Solowjow, C. Wen, J. Aparicio, A.M. Agogino, “Deep reinforcement learning for robotic assembly of mixed deformable and rigid objects” in  IEEE/RSJ International Conference on Intelligent Robots and Systems (IROS) pp. 2062-2069, 2018.

\bibitem{johannink19} T. Johannink, S. Bahl, A. Nair, J. Luo, A. Kumar, M. Loskyll, J. Aparicio, E. Solowjow, S. Levine, “Residual reinforcement learning for robot control.” in IEEE international conference on robotics and automation (ICRA) pp. 6023-6029, 2019.

\bibitem{c0} C. Eppner, S. Höfer, R. Jonschkowski, R. Martín-Martín, A. Sieverling, V. Wall, and O. Brock, “Lessons from the amazon picking challenge: four aspects of building robotic systems.” in Robotics: Science and Systems, 2016.

\bibitem{c1} C. Hernandez, M. Bharatheesha, W. Ko, H. Gaiser, J. Tan, K. van Deurzen, M. de Vries, B. Van Mil, et al., “Team delft’s robot winner of the amazon picking challenge 2016,” in Robot World Cup. Springer, pp. 613–624, 2016.

\bibitem{c2} A. Zeng, K.-T. Yu, S. Song, D. Suo, E. Walker, A. Rodriguez, and J. Xiao, “Multi-view self-supervised deep learning for 6d pose estimation in the amazon picking challenge,” in IEEE international conference on robotics and automation (ICRA) pp. 1386–1383, 2017.

\bibitem{c3} M. Nieuwenhuisen, D. Droeschel, D. Holz, J. Stückler, A. Berner, J. Li,R. Klein, and S. Behnke, “Mobile bin picking with an anthropomorphic service robot,” in 2013 IEEE International Conference on Robotics and Automation  pp. 2327–2334, 2013.

\bibitem{c4} M.-Y. Liu, O. Tuzel, A. Veeraraghavan, Y. Taguchi, T. K. Marks, and R. Chellappa, “Fast object localization and pose estimation in heavy clutter for robotic bin picking,” The International Journal of Robotics Research, 31(8), pp. 951–973, 2012.

\bibitem{c5} Lenz, H. Lee, and A. Saxena, “Deep learning for detecting robotic grasps,” The International Journal of Robotics Research, 34(4), pp. 705–724, 2015.

\bibitem{c6} L. Pinto, J. Davidson, and A. Gupta, “Supervision via competition: Robot adversaries for learning tasks,” in IEEE International Conference on Robotics and Automation (ICRA), pp. 1601–1608, 2017.

\bibitem{Bohg_c7_0}
J. Bohg, A. Morales, T. Asfour, and D. Kragic (2013). "Data-driven grasp synthesis—a survey". IEEE Transactions on Robotics, 30(2), pp. 289-309, 2013.

\bibitem{Kappler_c7_1}
D. Kappler, J. Bohg, and S. Schaal, "Leveraging big data for grasp planning". In IEEE International Conference on Robotics and Automation (ICRA), pp. 4304-4311, 2015.

\bibitem{Morrison_c7_2}
D. Morrison, P. Corke, and J. Leitner, "Closing the loop for robotic grasping: A real-time, generative grasp synthesis approach." In Robotics: Science and Systems, 2018.

\bibitem{Mahler17_c7_3}
J. Mahler et al. "Dex-Net 2.0: Deep Learning to Plan Robust Grasps with Synthetic Point Clouds and Analytic Grasp Metrics.", In Robotics: Science and Systems (RSS), 2017.

\bibitem{c7} J. Mahler, M. Matl, V. Satish, M. Danielczuk, B. DeRose, S. McKinley, and K. Goldberg, “Learning ambidextrous robot grasping policies,” Science Robotics, 4(26), 2019.

\bibitem{c8} FANUC, "Vision Functions for robots", https://www.fanuc.eu/pt/en/robots/accessories/robot-vision, Website and Product brochure, Accessed on 22 Feb. 2020.

\bibitem{c9} KUKA, "KUKA.PLC mxAutomation", www.kuka.com/en-us/products/robotics-systems/software/hub-technologies/kuka,-d-,plc-mxautomation, Website, Accessed on 24 Feb. 2020.

\bibitem{c10} Siemens, "SIMATIC S7-1500/ET 200MP Automation System in a Nutshell", siemens.com, Oct. 2016.

\bibitem{inpaint_c11} A. Telea, "An image inpainting technique based on the fast marching method", in Journal of graphics tools 9(1), Taylor \& Francis, pp. 23--34, 2004.

\bibitem{Satish19_c13}
V. Satish, J. Mahler and K. Goldberg, "On-policy dataset synthesis for learning robot grasping policies using fully convolutional deep networks", in IEEE Robotics and Automation Letters, 4(2), pp. 1357--1364, 2019.

\bibitem{danielczuk17_c12} M. Danielczuk, M. Matl, S. Gupta, A. Li, A. Lee, J. Mahler, K. Goldberg, "Segmenting unknown 3d objects from real depth images using mask r-cnn trained on synthetic data", IEEE  International Conference on Robotics and Automation (ICRA), pp. 7283--7290, 2019.








\end{thebibliography}
\end{document}